\pdfoutput=1
\documentclass[11pt]{article}

\usepackage{microtype}
\usepackage{booktabs}
\usepackage{url}
\usepackage{xurl}    
\usepackage{csquotes}

\usepackage{amsmath}
\usepackage{amssymb}
\usepackage{mathtools}
\usepackage{amsthm}

\usepackage{graphicx}
\usepackage{listings}
\usepackage{float}
\usepackage{xcolor}

\usepackage[margin=1in]{geometry}
\usepackage{authblk}

\usepackage[subrefformat=parens]{subcaption}
\usepackage{hyperref}
\usepackage{natbib}
\usepackage{worldflags} 

\theoremstyle{plain}
\newtheorem{theorem}{Theorem}[section]

\theoremstyle{definition}
\newtheorem{definition}[theorem]{Definition}

\theoremstyle{remark}

\graphicspath{{./figures/specific_problem/}}

\definecolor{codegreen}{rgb}{0,0.6,0}
\definecolor{codegray}{rgb}{0.5,0.5,0.5}
\definecolor{codepurple}{rgb}{0.58,0,0.82}
\definecolor{backcolour}{rgb}{0.95,0.95,0.92}
\lstdefinestyle{pythonstyle}{
    backgroundcolor=\color{backcolour},
    commentstyle=\color{codegreen},
    keywordstyle=\color{magenta},
    numberstyle=\tiny\color{codegray},
    stringstyle=\color{codepurple},
    basicstyle=\ttfamily\footnotesize,
    breakatwhitespace=false,
    breaklines=true,
    captionpos=b,
    keepspaces=true,
    numbers=left,
    numbersep=5pt,
    showspaces=false,
    showstringspaces=false,
    showtabs=false,
    tabsize=2
}
\title{Synthesizing Feature Extractors: \\An Agentic Approach for Algorithm Selection\thanks{Code, data, and the full reproducibility archive: \url{https://doi.org/10.5281/zenodo.20161743}.}}

\author[1]{Hai Xia}
\author[2]{Carlos Ansótegui}
\author[1]{Stefan Szeider}

\affil[1]{Algorithms and Complexity Group, TU Wien, Vienna, Austria\\
  \texttt{\{hxia,sz\}@ac.tuwien.ac.at}}
\affil[2]{Logic \& Optimization Group, University of Lleida, Lleida, Spain\\
  \texttt{carlos.ansotegui@udl.cat}}
\date{}

\hypersetup{%
  pdfauthor={Hai Xia, Carlos Ansótegui, Stefan Szeider},
  pdftitle={Synthesizing Feature Extractors: An Agentic Approach for Algorithm Selection},
  pdfsubject={Algorithm selection via LLM-synthesized feature extractors},
  pdfkeywords={algorithm selection, large language models, feature extraction, constraint optimization, MiniZinc}
}

\begin{document}

\maketitle

\begin{abstract}
Algorithm selection for constraint satisfaction problems requires extracting
features that capture problem structure. Manually designing feature extractors
demands deep domain expertise and quickly becomes a bottleneck when
new problem classes appear. We present an automated approach that uses Large
Language Models (LLMs) in an agentic check--fix--verify loop to synthesize
executable Python scripts that act as interpretable, problem-specific
feature extractors. Given a high-level MiniZinc model and an instance, the LLM
agent generates code that constructs a typed graph representation and
computes structural properties such as graph density, variable clustering,
and constraint tightness. We evaluate our approach on three combinatorial
problems (vehicle routing, car sequencing, fixed-length error-correcting
codes) with a portfolio of five state-of-the-art solvers. The synthesized
extractors yield algorithm selectors that consistently outperform both
expert-curated \emph{mzn2feat} features (up to $8.3$~percentage points~(pp)
test-set accuracy on FLECC) and the best transformer-based \emph{trans2feat}
variants. In the meanwhile, the synthesized feature extractors  remain inspectable.
\end{abstract}

\section{Introduction}
\label{sec:intro}

Combinatorial problems such as vehicle routing (VRP), car sequencing (CS), and
fixed-length error-correcting codes (FLECC) are computationally hard:
state-of-the-art solvers can run for hours without finding proven optima, and
the gap between the best and a poorly chosen solver on an instance can span
orders of magnitude~\citep{kotthoff2016algorithm,kershke19automated}.
The difficulty of an instance depends strongly on its structure (customer
density, time-window tightness, constraint coupling), and that structure
determines which solver performs
best~\citep{smithmiles2012measuring}. Algorithm selection (AS) exploits this
fact by mapping instances to the most effective solvers from a portfolio
of complementary solvers~\citep{rice1975algorithm,kershke19automated}.

The standard AS pipeline depends on a \emph{feature extractor} that computes an informative feature vector from the
problem instance. However, designing such
an extractor typically requires substantial domain expertise and insight into
which structural properties correlate with solver performance. Moreover, validating an extractor's efficiency and effectiveness across instances takes a long time. Therefore, AS has been applied primarily to domains with mature formalisms, chiefly SAT~\citep{hoos2021automated,shavit2024revisit} and
constraint programming via
\emph{mzn2feat}~\citep{amadini13Features,amadini14enhanced}. When it comes to problem
classes outside classical formalisms, researchers either build a new extractor
from scratch or translate the problem into one of them, with possible structural information loss.

In general, there are several sides preventing the AS pipeline from being widely used. On the design side, it requires
expert-curated extractors like \emph{mzn2feat}. But for these extractors, it is still possible to miss solver-relevant
structural properties that our experiments
expose (Section~\ref{sec:experiments}). The expertise-necessary design also makes feature extractors generated without full autonomy. However, LLMs (Large Language Models) are trained with diverse domain knowledge, making them suitable to curate feature extractors. On the LLM side, a naive single-shot prompt does not work in practice:
weaker open-source backends fail to produce a working extractor in $10/10$ trials, and removing any step in our loop below
collapses success to zero even for strong models
(Sections~\ref{sec:framework} and~\ref{sec:cost}).
The technical question is how to wrap an LLM so its outputs are
reliably executable, AS-relevant, and inspectable.

\paragraph{Our approach: automating feature extraction via LLMs}
We introduce an LLM-based framework that automatically generates
executable Python scripts as feature extractors. The design is a
\emph{two-level process}: an LLM agent, wrapped in a
check--fix--verify error-correcting loop, synthesizes the program,
and the output program is then the extractor. The agent first reads a
high-level MiniZinc problem
description~\citep{stuckey14minizinc,marriott2008design} and produces a
Python script. Then the extractor (the Python script) constructs a graph
representation from an instance and outputs a vector of interpretable features. As MiniZinc is
declarative formalisms with rich information, like give the LLM a compact view of the problem's structural patterns, we use MiniZinc as our problem modeling environment.

The framework produces explicit feature extractors (Python programs), and the output features are interpretable rather than opaque
neural embeddings. While recent work has explored deep learning for producing
latent problem representations~\citep{pellegrino2025transformer,zhang2024grass,loreiggia2016deep},
such approaches sacrifice transparency for automation. Our generated
extractors produce graphical features, like graph density, variable clustering, constraint
tightness, statistical summaries of data that domain experts can read, validate, and refine. This ``gray-box'' design keeps the automation accessible to human understanding and improvement.

\paragraph{Empirical validation}
We validate the approach on AS for three combinatorial problems (VRP, CS,
FLECC) using a portfolio of five state-of-the-art solvers (Gurobi, CPLEX,
SCIP, Gecode, OR-Tools). The synthesized extractors outperform both
\emph{mzn2feat}, the established expert-curated extractor for MiniZinc
problems~\citep{amadini13Features,amadini14enhanced}, and the
transformer-based \emph{trans2feat}~\citep{pellegrino2025transformer}.
The gain comes from capturing high-level structural properties that
flat representations and opaque embeddings miss.

\paragraph{Contributions}
\begin{enumerate}
\item We demonstrate that an LLM agent can reason about combinatorial problem structure and synthesize functional, interpretable feature
extractors from MiniZinc problem descriptions, reducing the manual
engineering cost of building AS pipelines for problems expressible in
MiniZinc.
\item We propose an agentic check--fix--verify pipeline whose
intermediate artifact is an explicit Python program. Unlike opaque neural embeddings, the generated extractors expose inspectable features.
\item Across VRP, CS, and FLECC, selectors built from our synthesized
features outperform selectors built from expert-curated \emph{mzn2feat} and
transformer-based \emph{trans2feat}. This suggests the LLM uncovers
solver-aware structural patterns that expert-curated extractors and transformer-based pipelines miss.
\end{enumerate}

\section{Related Work}
\label{sec:related}

The \emph{Algorithm Selection Problem}
(AS)~\citep{rice1975algorithm,kershke19automated}
considers a portfolio $\mathcal{P}$ of algorithms, a set of instances $I$, a
performance metric $PM(A,i)$, and a resource budget $B$. Since the
performance of an algorithm $A \in \mathcal{P}$ varies across instances, an
AS strategy must predict, \emph{before solving}, which $A$ to run on a given
instance. To make this tractable, each instance $i \in I$ is described by a
\emph{feature vector} $\phi(i) \in \mathbb{R}^d$ obtained from a feature
extractor $\Phi: i \mapsto \phi(i)$. The AS task is to learn a selector
$S:\mathbb{R}^d \rightarrow \mathcal{P}$ maximizing
$\sum_{i \in I} PM(S(\phi(i)), i)$ subject to $B$. Two reference points calibrate AS performance:

\begin{definition}[Single Best Solver]
\label{def:sb}
The \emph{Single Best Solver} (SB) is the algorithm
$A^{\text{SB}} = \arg\max_{A \in \mathcal{P}} PM(A, I)$ that achieves the
best overall performance across the entire instance set $I$. The SB
strategy applies $A^{\text{SB}}$ to every instance.
\end{definition}

\begin{definition}[Virtual Best Solver]
\label{def:vbs}
The \emph{Virtual Best Solver} (VBS) is the (hypothetical) per-instance
selector that, for each $i \in I$, chooses the algorithm
$A \in \mathcal{P}$ that achieves the best performance on that instance:
$PM(\text{VBS}, I) = \sum_{i \in I} \max_{A \in \mathcal{P}} PM(A, \{i\})$.
The VBS upper-bounds the performance of any AS strategy.
\end{definition}

MiniZinc~\citep{nethercote2007minizinc} is a high-level, declarative modeling
language for constraint satisfaction and optimization. A model file
(\texttt{.mzn}) defines variables, constraints, and (optionally) an objective. A data file (\texttt{.dzn}) contains instance parameters.

\paragraph{Expert-curated features}
Classical AS builds on hand-engineered feature sets such as
SATzilla's~\citep{shavit2024revisit} and
\emph{mzn2feat}'s~\citep{amadini13Features,amadini14enhanced}; hand-crafted
features likewise drive per-instance policy selection inside solvers, e.g.\
in SAT-based tree decomposition~\citep{xia24sat}.

\paragraph{Graph-based features without LLMs}
Encoding combinatorial instances as graphs for feature extraction does not by
itself require an LLM. \citet{stone2024domain} convert instances from three
problem domains into domain-agnostic graph and image encodings, extract
generic graph metrics, and use the resulting representations for algorithm
selection and other downstream tasks. Their pipeline applies one fixed,
hand-specified encoding and feature set across all domains; our agent instead
\emph{writes a new extractor program per problem family}, deciding from the
high-level MiniZinc model which typed graph to construct and which
problem-adapted, solver-aware quantities to materialize as named, editable
features (e.g.\ demand concentration and depot eccentricity for VRP,
Section~\ref{sec:qualitative}). The approaches are complementary: generic
encodings transfer at zero synthesis cost, synthesized extractors capture
semantics that fixed encodings discard.

\paragraph{LLM-based feature engineering and program synthesis}
CAAFE~\citep{hollmann2023caafe} uses LLMs to generate Python feature
transformations for tabular ML datasets, and
FeatLLM~\citep{han2024featllm} prompts LLMs to derive rule-based features
for few-shot tabular learning. Both operate on existing tabular inputs. More
broadly, LLM-driven program search can yield executable artifacts that are
competitive in combinatorial settings~\citep{romera2024funsearch}. Our work
instead starts from declarative \texttt{.mzn}/\texttt{.dzn} specifications
and synthesizes a complete, reusable extractor program that computes
structural and semantic properties of the problem specification itself.

\paragraph{Neural embeddings for algorithm selection}
\citet{wu2024asllm} use LLMs to embed algorithm source code and
documentation for AS. \citet{zhang2024grass} combine graph neural networks
with expert knowledge to select SAT solvers, learning embeddings of CNF
formula structure. \citet{pellegrino2025transformer} apply transformer
encoders directly to the high-level textual representation of constraint
optimization instances to learn features. These neural
approaches produce high-dimensional embeddings whose individual dimensions
lack clear semantic meaning and cannot easily be inspected or edited. On
the FLECC and CS instance sets for which~\citet{pellegrino2025transformer}
released \emph{trans2feat} features, our \emph{LLM2feat} selectors
outperform the best \emph{trans2feat} variant by $7.4$ and $5.4$
percentage points~(pp) in test accuracy
(Section~\ref{sec:trans2feat-comparison}).

\section{Problem-Specific LLM-Based Agent}
\label{sec:framework}

We build on the agentic framework that bridges LLMs and constraint
satisfaction solving~\citep{szeider2025briding}.

\begin{definition}[LLM agent]
\label{def:llm-agent}
A \emph{Large Language Model agent}~\citep{yao2023react} is a tuple
$\mathcal{A} = (L, T, M, \pi, E)$, where $L$ is a language model, $T$ is
a set of external tools, $M$ is a memory module, $\pi$ is the prompting
policy that maps observations and history to model inputs, and $E$ is
the environment. The agent operates in a loop
$o_t \xrightarrow{\pi} p_t \xrightarrow{L} a_t \xrightarrow{T, E} o_{t+1}$,
where $o_t$ is the observation at time $t$, $p_t$ is the constructed
prompt, $a_t$ is the action (e.g.\ tool call), and $o_{t+1}$ is the
next observation.
\end{definition}

\begin{figure}[htbp]
  \centering
  \includegraphics[trim=0 0 0 0,width=0.95\linewidth]{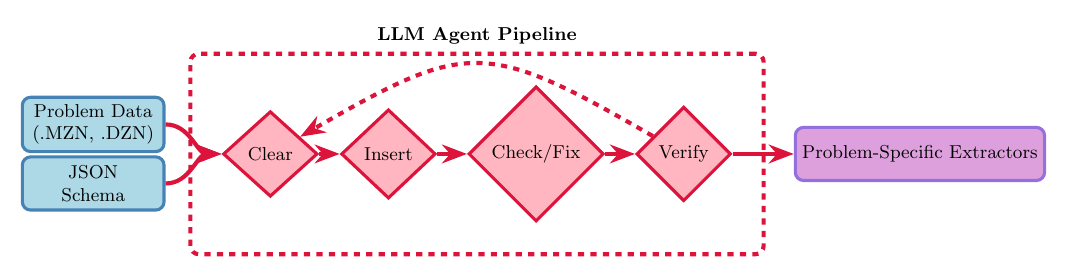}
  \caption{Workflow for generating problem-specific feature extractors. The
  agent loops over \textbf{Clear}, \textbf{Insert}, \textbf{Check/Fix}, and
  \textbf{Execute} steps until an executable Python script that produces a
  validated feature vector is obtained.}
  \label{fig:problem_specific_framework}
\end{figure}

Our agent takes a MiniZinc problem description (\texttt{.mzn} and
\texttt{.dzn} files) and a data schema as input, and outputs a Python
script that extracts a feature vector from any instance of that
problem. Two prompts control the loop in
Figure~\ref{fig:problem_specific_framework}:
\emph{script\_system\_prompt} (a general-purpose Python-script
generation prompt that defines a strict workflow, technical
requirements, and the available tools) and \emph{mzn-tuning} (a
domain-specialized prompt that instructs the agent to extract $50$ standardized structural features from constraint-programming problems
suitable for AS).

\paragraph{General script prompt}
The general Python-script prompt (\texttt{script-system-prompt.md})
enforces a four-step workflow:
(i)~\textbf{Clear} all previous content;
(ii)~\textbf{Insert} the complete script;
(iii)~\textbf{Check/Fix} syntax and structural requirements, addressing
validation errors if needed;
(iv)~\textbf{Execute} the script and validate its output. The expected
script structure is given in Listing~\ref{lst:general}.

\begin{lstlisting}[language=Python, caption={General Python script template
guided by \texttt{script-system-prompt.md}.}, label=lst:general, float=htbp]
# Required script structure
import necessary_modules
# Constants and configuration
CONSTANTS = values
# Function definitions
def helper_functions():
    pass
# Main execution logic
if __name__ == "__main__":
    # Processing logic here
    result_dict = {"key": "value", "results": data}
# MANDATORY: Output results
output_results(result_dict)  # Must be final line
\end{lstlisting}

\paragraph{MiniZinc tuning prompt}
The specialized \emph{mzn-tuning} prompt (\texttt{mzn-tuning-prompt.md})
layers AS-specific requirements on top of the general workflow: mandatory
imports for framework integration, no file I/O (instance data is
accessed via \texttt{input\_data()}), and a standardized output shape
of exactly $50$ features ready for downstream selector training. The
template is given in Listing~\ref{lst:mzntuning}. The size $50$ was fixed
in pilot experiments as a balance between coverage and interpretability:
comparable in scale to \emph{mzn2feat}'s $95$ and \emph{trans2feat}'s $116$
dimensions, yet small enough that each feature remains individually
inspectable and selectors stay data-efficient. The \emph{identity} of the
$50$ features is chosen anew by the agent in every run: across the $10$
independent syntheses per problem
(Section~\ref{sec:trans2feat-comparison}), core structural metrics such as
graph density, degree statistics, and clustering recur consistently, while
the remaining features vary in which problem-specific quantities they
materialize (Section~\ref{sec:qualitative}); the per-run feature lists and
value tables ship with the supplement. Each step of the
check--fix--verify loop is load-bearing: in pilot ablations, removing
any one of the four steps produced zero executable extractors, and
weaker LLM backends that drift from the protocol fail in $10/10$
trials (Section~\ref{sec:cost}).

\begin{lstlisting}[language=Python, caption={MiniZinc instance feature
extraction template guided by \texttt{mzn-tuning-prompt.md}.},
label=lst:mzntuning, float=htbp]
# MANDATORY imports (exact format required)
from lmtune_helpers import input_data, output_results
import networkx as nx
import numpy as np
def main():
    # Get instance data (no file I/O allowed)
    instance_data = input_data()
    # Initialize standardized results structure
    results = {
        "README": "~200 word methodology description",
        "characteristic_1": 0.0,  # Problem size metrics
        "characteristic_2": 0.0,  # Graph properties
        # ... (extract exactly 50 characteristics)
        "characteristic_50": 0.0  # Structural complexity
    }
    # Analyze constraint optimization instance
    # Extract solver-relevant characteristics:
    # - Problem size (variables, constraints)
    # - Graph properties (density, clustering, centrality)
    # - Data distribution (statistical properties)
    # - Structural complexity (symmetries, sparsity)
    # MANDATORY: Return standardized results
    output_results(results)
if __name__ == "__main__":
    main()
\end{lstlisting}

\section{Experimental Analyses}
\label{sec:experiments}

We evaluate the agentic framework on three problem-specific AS
benchmarks and analyze \emph{why} \emph{LLM2feat} features yield better
selectors than \emph{mzn2feat} or \emph{trans2feat}, looking at feature
correlation, utilization efficiency, and selector accuracy.

\subsection{Experimental Settings}
\label{sec:settings}

We use the top-performing solvers from the
MiniZinc challenge\footnote{\url{https://www.minizinc.org/challenge/2025/results/}}:
Gurobi (12.0.3), CPLEX (22.1.2), SCIP (9.2.3), Gecode (6.2.0), and OR-Tools
(9.3.10497). The portfolio spans solver paradigms: Gurobi and CPLEX are
commercial mixed-integer programming (MIP) solvers; SCIP combines CP and
MIP techniques; Gecode is a constraint-programming (CP) solver used
throughout the MiniZinc Challenge history; OR-Tools won gold medals across
all major categories of the MiniZinc Challenge in 2023--2025.

The instance set $I$ contains minimization, maximization, and decision
problems: ``best solver'' means the lowest/highest objective value at
timeout for optimization problems, and the shortest solving time for
decision problems; the performance metric is oriented so that larger
values are better. We use three problems with sufficient instance
diversity: VRP~\citep{queiroga2022optiaml}, CS~\citep{pellegrino2025transformer},
and FLECC~\citep{pellegrino2025transformer}. Each problem is split $7{:}3$
into training/test sets; the same split is reused across all extractor
comparisons. Solvers run with a $20$-minute timeout per instance, in line
with the MiniZinc Challenge standard\footnote{\url{https://www.minizinc.org/challenge/}};
extractor synthesis is capped at $1$ minute per attempt, with automatic
restart on timeout. All experiments run on a
cluster with two AMD~7403 processors ($24$ cores at 2.8\,GHz, $32$\,GB
RAM/core). Two performance metrics are reported: $Acc$, the fraction of
instances on which the selector picks the truly best solver, and $Rank$,
the average ranking of the selected solver (lower is better).

For LLM model selection, we use the commercial OpenAI o4-mini-2025-04-16
as the agent backend. All LLM calls use the provider's default generation
parameters: the sampling temperature is never overridden (o4-mini is a
reasoning model whose temperature is fixed at its default of $1.0$), and
the reasoning effort is the API default (\texttt{medium}); the exact call
sites are in the supplement (\texttt{llm\_factory.py}). Run-to-run
stochasticity is thus a property of the backend itself and is quantified
through $10$ independent synthesis runs per problem. The open-weight
backends we evaluated (Llama~3.3, DeepSeek~R1) failed to follow the
check--fix--verify protocol in our setup (Section~\ref{sec:cost});
sensitivity analyses across LLM variants and the $10$ runs appear in
Appendix~\ref{appendix:llm-sensitivity}.

\subsection{Algorithm-Selection Toolchains}
\label{sec:toolchains}

The AS input is a feature table (instance $\rightarrow$ feature vector) and
a performance table (instance $\rightarrow$ per-solver performance). We
compare three feature extractors: our LLM-synthesized \emph{LLM2feat},
expert-curated \emph{mzn2feat}~\citep{amadini13Features,amadini14enhanced},
and transformer-based
\emph{trans2feat}~\citep{pellegrino2025transformer}.

For selector training we use Random Forest (RF) following the standard
recipe~\citep{kershke19automated} and
AutoSklearn (AutoSK)~\citep{feurer15efficient,feurer20Autosklearn};
hyperparameters are in Appendix~\ref{appendix:hyperparams}. We also report
AutoFolio (AF)~\citep{lindauer15autofolio} and
LLAMA~\citep{kotthoff2013llama}, top-performing tools in AS surveys and
challenges~\citep{kershke19automated}. RF and AutoSK are trained with two loss functions, $Acc$
and $Rank$; AF and LLAMA use their default $Acc$ loss. All training uses
5-fold cross-validation.

\subsection{Research Questions}

\textbf{Q1:} How diverse are the features generated by \emph{LLM2feat}?
Highly correlated features carry overlapping information and inflate the
effective dimension of the feature space.
\textbf{Q2:} How efficiently does each feature contribute to the AS
model? Distributions over feature importance shed light on quality and
potential redundancy.
\textbf{Q3:} How accurate are the resulting AS models compared with \emph{mzn2feat}-based and \emph{trans2feat}-based selectors?

\subsection{Feature Correlation Analysis (Q1)}
\label{sec:correlation}

\begin{figure}[htbp]
  \centering
  \includegraphics[width=0.55\linewidth]{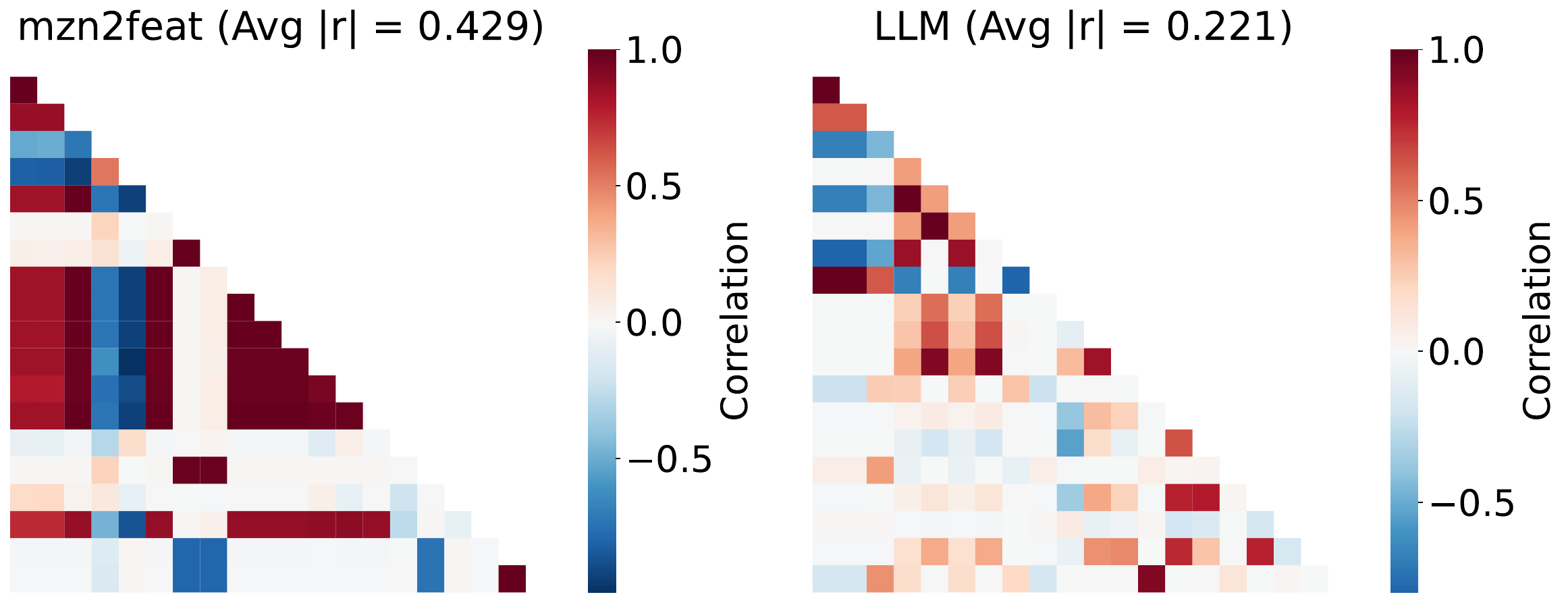}
  \caption{Feature correlation matrices for VRP (feature names suppressed
  for clarity). \emph{LLM2feat} features show $48.5\%$ lower average
  correlation ($|r|=0.221$) than \emph{mzn2feat} ($|r|=0.429$).}
  \label{fig:vrp_correlation}
\end{figure}

For each feature set we train an RF selector as in
Section~\ref{sec:settings}, extract the top $20$ features by importance
for the random forest model, and compute Pearson correlation
matrices~\citep{guyon2003introduction} on the selected features.
Figure~\ref{fig:vrp_correlation} shows the result for VRP: \emph{LLM2feat}
features have a $48.5\%$ lower average absolute correlation
($|r|=0.221$) than \emph{mzn2feat} ($|r|=0.429$). The same pattern holds
for FLECC ($|r|=0.306$ vs.\ $0.330$) and CS ($|r|=0.551$ vs.\ $0.725$).
The corresponding heatmaps for FLECC and CS are in
Appendix~\ref{appendix:correlation-detail}. LLM-generated features
capture a more diverse range of structural properties, reducing
redundancy with fewer overlapping features.

\subsection{Feature Utilization Efficiency (Q2)}
\label{sec:utilization}

\begin{figure}[htbp]
  \centering
  \begin{subfigure}[t]{0.32\linewidth}
    \centering
    \includegraphics[width=\linewidth]{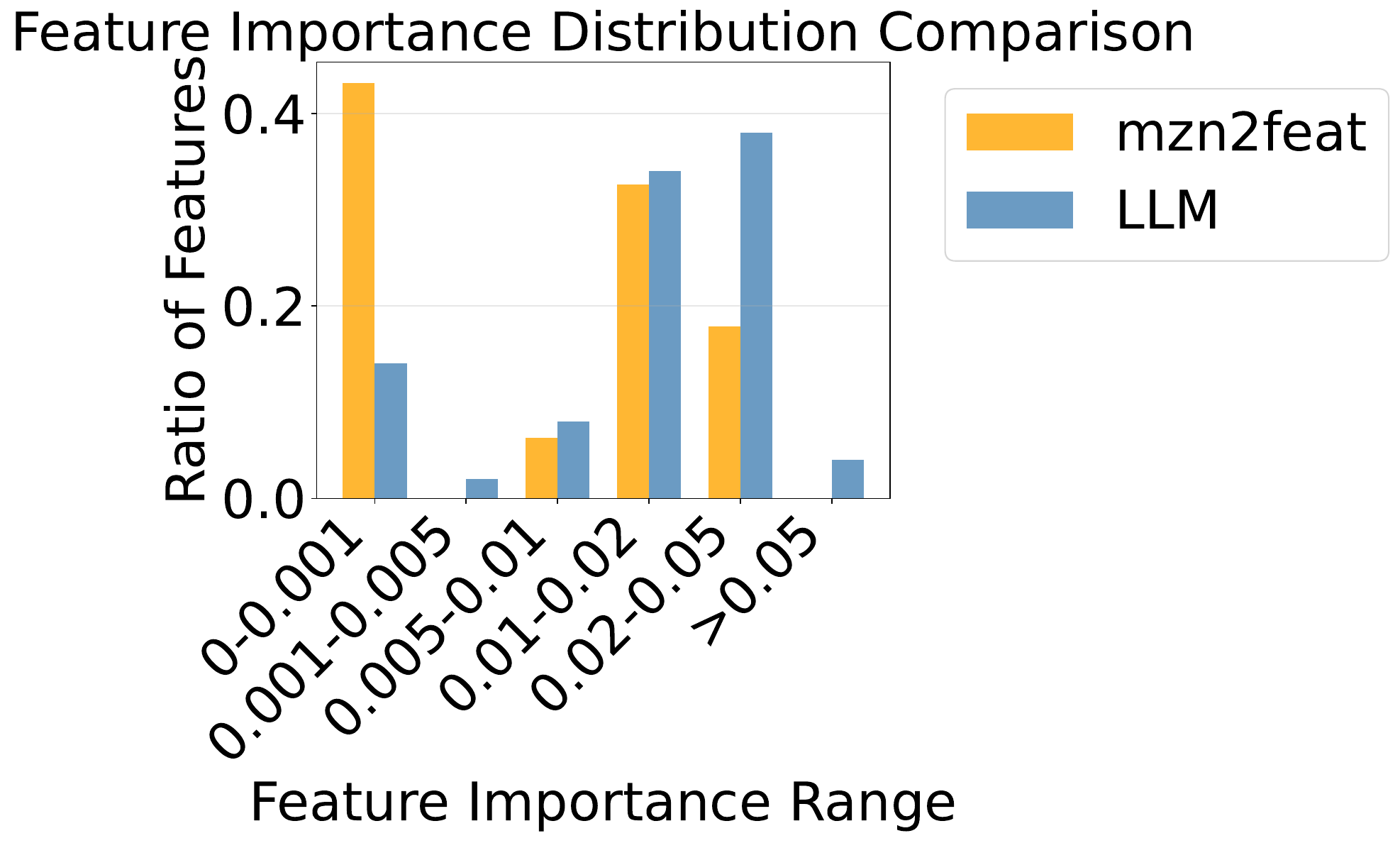}
    \caption{FLECC}
  \end{subfigure}
  \begin{subfigure}[t]{0.32\linewidth}
    \centering
    \includegraphics[width=\linewidth]{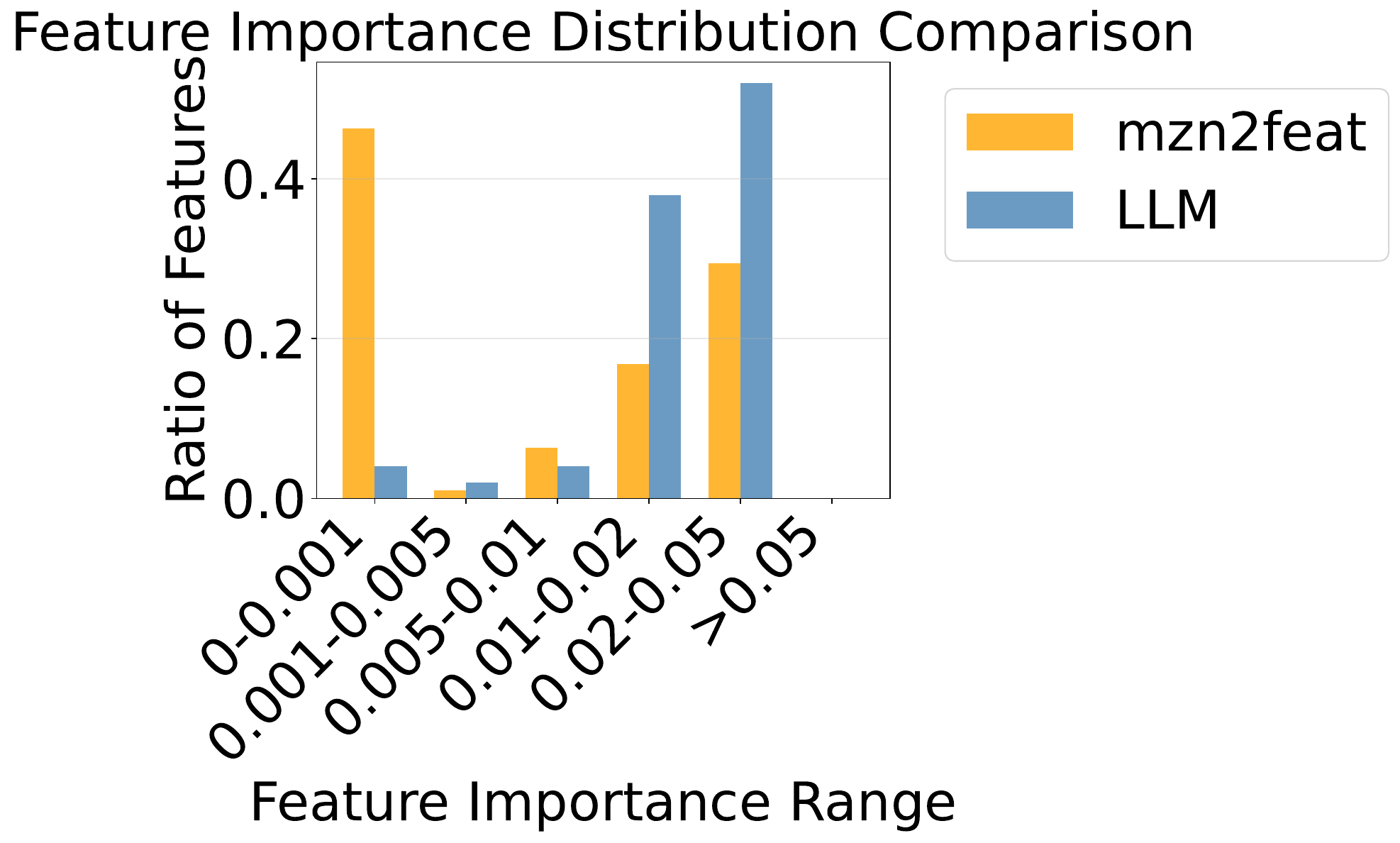}
    \caption{CS}
  \end{subfigure}
  \begin{subfigure}[t]{0.32\linewidth}
    \centering
    \includegraphics[width=\linewidth]{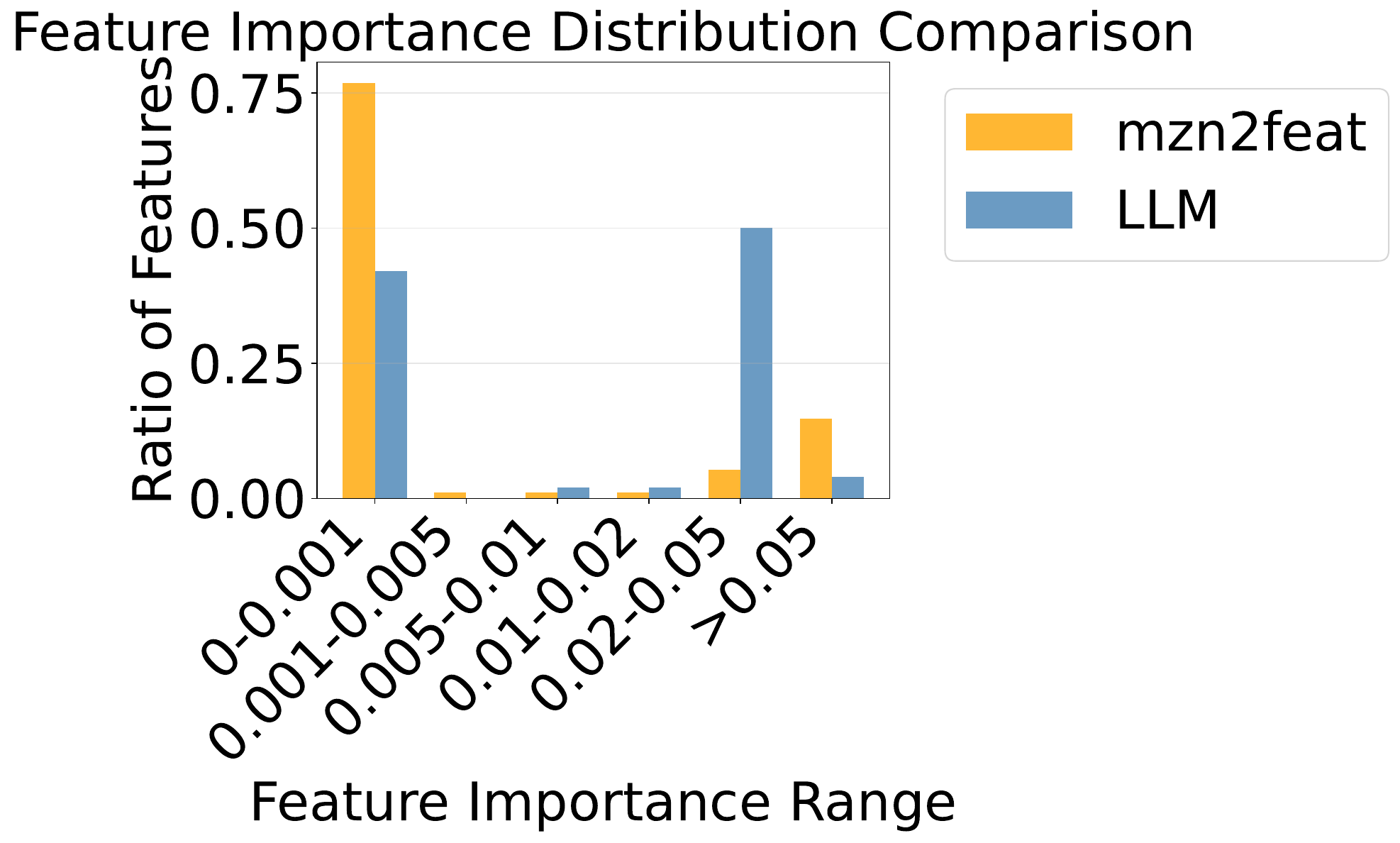}
    \caption{VRP}
  \end{subfigure}
  \caption{Feature-importance distribution. \emph{LLM2feat} achieves
  better spread of importance across features than \emph{mzn2feat} on
  all three problems.}
  \label{fig:cross_problem_distribution}
\end{figure}

For both \emph{mzn2feat}+RF and \emph{LLM2feat}+RF we extract
feature-importance scores for the entire feature set and count features
above a $0.001$ significance threshold as ``effectively utilized.''
\emph{mzn2feat} provides $95$ hand-crafted features; \emph{LLM2feat}
produces $50$. Across the three problems, \emph{LLM2feat} achieves significantly higher utilization
(Figure~\ref{fig:cross_problem_distribution}): on VRP, $96\%$ vs.\
$56.8\%$ (a $69\%$ relative improvement); on FLECC, $58\%$ vs.\
$23.2\%$; on CS, $84\%$ vs.\ $45.1\%$. More of the LLM-generated
dimensions contribute meaningfully to the selector's decisions.

\subsection{Accuracy Analysis (Q3)}
\label{sec:accuracy}

\begin{figure}[htbp]
  \centering
  \begin{subfigure}[t]{0.32\linewidth}
    \centering
    \includegraphics[width=\linewidth]{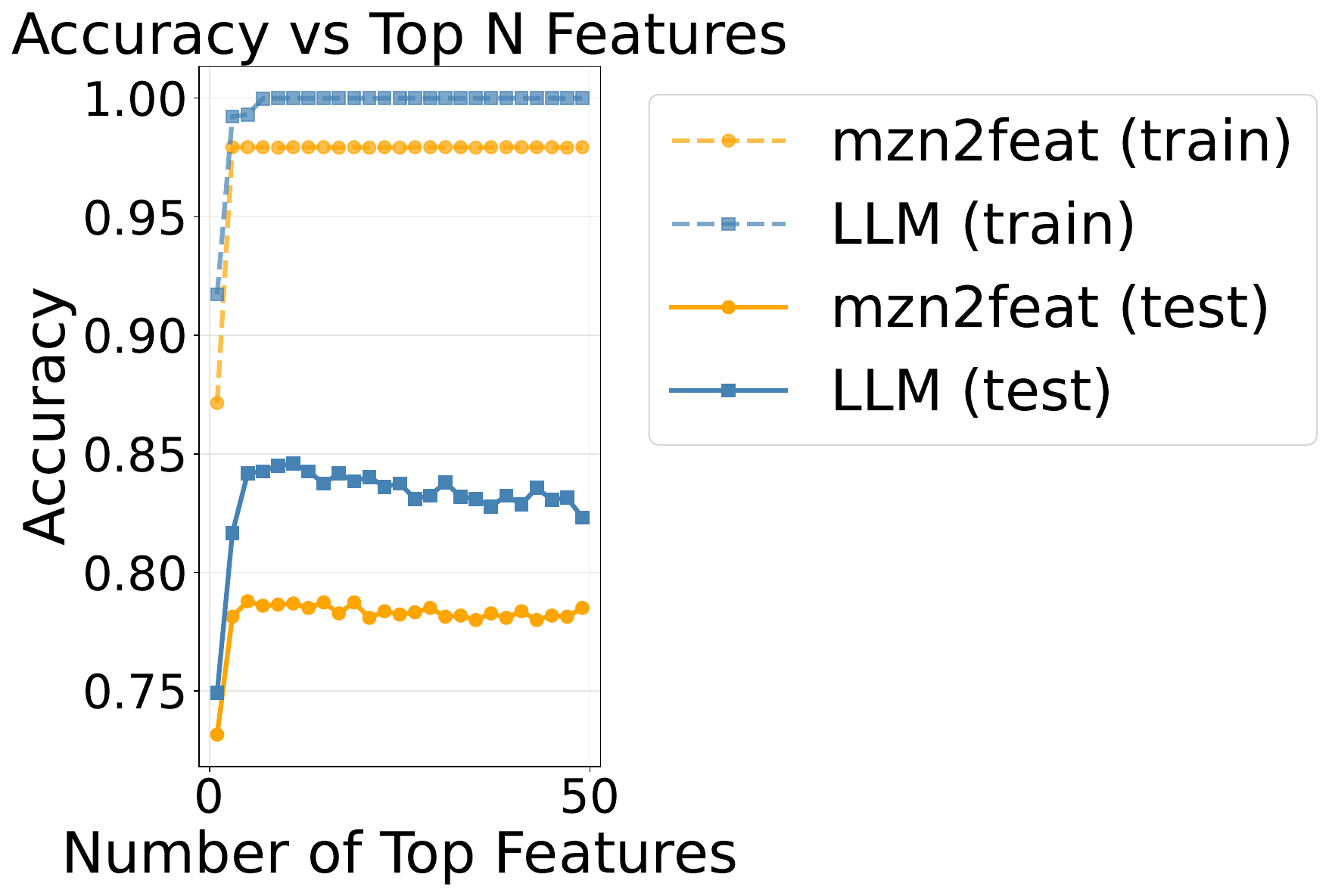}
    \caption{FLECC}
  \end{subfigure}
  \begin{subfigure}[t]{0.32\linewidth}
    \centering
    \includegraphics[width=\linewidth]{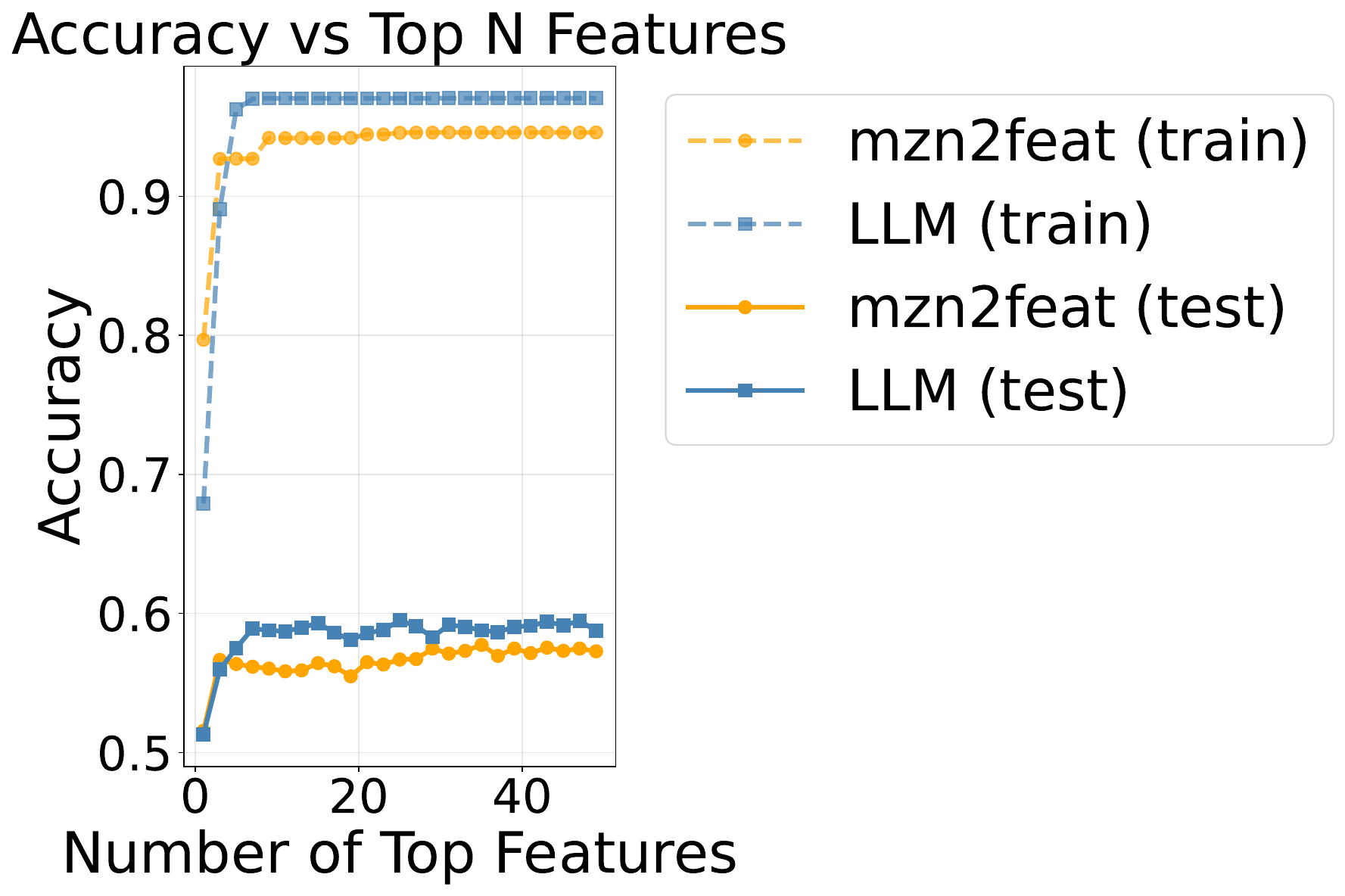}
    \caption{CS}
  \end{subfigure}
  \begin{subfigure}[t]{0.32\linewidth}
    \centering
    \includegraphics[width=\linewidth]{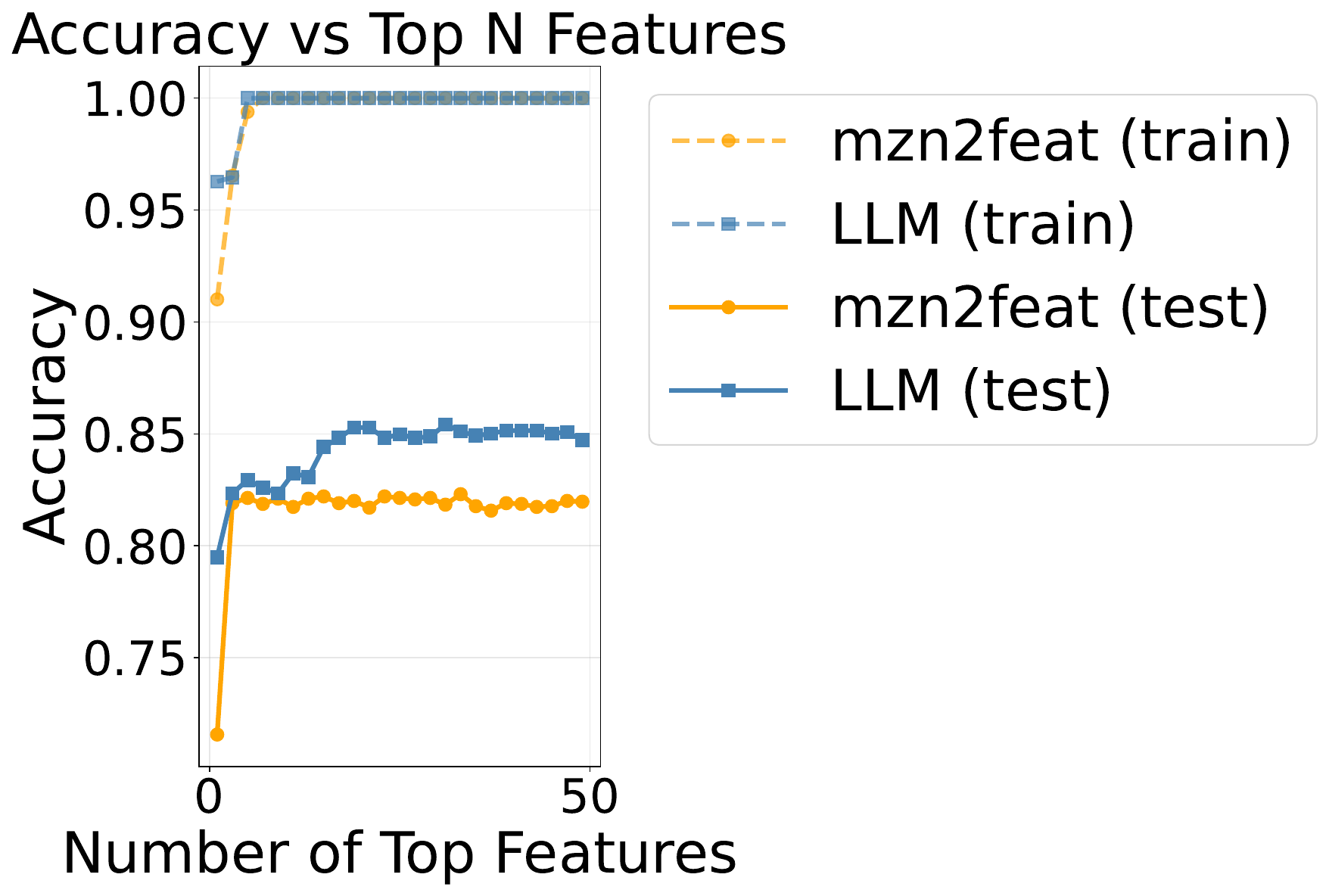}
    \caption{VRP}
  \end{subfigure}
  \caption{AS test accuracy as a function of feature-set size on the
  three benchmarks. \emph{LLM2feat} matches or exceeds \emph{mzn2feat}
  at every size and reaches higher peak accuracy.}
  \label{fig:problem_accuracy}
\end{figure}

We evaluate AS accuracy as a function of feature-set size, growing the
set in order of decreasing RF importance.
Figure~\ref{fig:problem_accuracy} shows that \emph{LLM2feat}-based
selectors reach higher accuracy with fewer features, and continue
to benefit from additional features where \emph{mzn2feat} plateaus: on
VRP, \emph{mzn2feat} flatlines near $81\%$ once $10$ features are used,
while \emph{LLM2feat} continues to improve and peaks beyond $20$ features.

Table~\ref{tab:accuracy_loss_acc_test} compares the three feature
extractors across four toolchains (AF, RF, LLAMA, AutoSK). With the
(trivial) exception of AF (which falls back to the single best solver,
SB, for all our extractors in this setting), \emph{LLM2feat} achieves
the highest accuracy and the lowest ranking on every problem--toolchain
pair. Across VRP, CS, and FLECC, \emph{LLM2feat} improves over the
\emph{mzn2feat} test accuracy by up to $8.3$\,pp, and over the best
\emph{trans2feat} variant by $7.4$\,pp on FLECC and $5.4$\,pp on CS
(see Section~\ref{sec:trans2feat-comparison} for the head-to-head with
all $20$ \emph{trans2feat} variants). Training-set results, ranking-loss
results, and full LLM-sensitivity tables are in
Appendix~\ref{appendix:llm-sensitivity}.

The three benchmarks include both classical AS regimes. On CS the single
best solver is weak ($49.0\%$), and feature-based selection adds up to
$15.0$\,pp. On VRP and FLECC the SB baseline
is strong ($79.5\%$/$78.2\%$), yet \emph{LLM2feat} still closes roughly
$30\%$ of the remaining gap between SB and the virtual best solver
($+6.3$\,pp on VRP, $+6.5$\,pp on FLECC over SB). Because the feature extraction and
selection add negligible overhead (less than 1 minute) relative to a $20$-minute solve, these gains convert directly into saved solver
time.

\begin{table}[htbp]
\caption{Test-set accuracy ($Acc$) and average ranking ($Rank$) for AS
toolchains using \emph{mzn2feat} ($95$~features), \emph{LLM2feat}
($50$~features), and \emph{trans2feat} ($116$~features). Loss
function~$=Acc$ (AutoFolio, LLAMA, and \emph{trans2feat} use their
accuracy-based defaults). Each \emph{LLM2feat} row reports the strongest
of the three initial synthesis runs per problem; best/mean/worst over all
$10$ runs appear in Table~\ref{tab:trans2feat_summary}. $\times$ marks
combinations unavailable because no released \emph{trans2feat} features
exist for VRP. Best per column in \textbf{bold}.}
\label{tab:accuracy_loss_acc_test}
\centering
\scalebox{0.9}{
\begin{tabular}{lcccccc}
\toprule
& \multicolumn{2}{c}{VRP} & \multicolumn{2}{c}{CS} & \multicolumn{2}{c}{FLECC} \\
\cmidrule(lr){2-3} \cmidrule(lr){4-5} \cmidrule(lr){6-7}
                  & $Acc$ & $Rank$ & $Acc$ & $Rank$ & $Acc$ & $Rank$ \\
\midrule
SB                & 79.5\% & 1.221 & 49.0\% & \textbf{1.616} & 78.2\% & 1.420 \\
\midrule
\emph{mzn2feat}+AF      & 79.5\% & 1.221 & 49.0\% & \textbf{1.616} & 78.2\% & 1.420 \\
\emph{mzn2feat}+RF      & 82.4\% & 1.195 & 54.7\% & 1.943 & 76.4\% & 1.426 \\
\emph{mzn2feat}+LLAMA  & 82.9\% & 1.184 & 59.4\% & 1.798 & 78.7\% & 1.397 \\
\emph{mzn2feat}+AutoSK  & 84.4\% & 1.166 & 62.1\% & 1.787 & 79.2\% & 1.396 \\
\midrule
\emph{trans2feat}+RF       & $\times$ & $\times$ & 52.9\% & 2.004 & 77.3\% & 1.447 \\
\emph{trans2feat}+AutoSK   & $\times$ & $\times$ & 54.5\% & 1.931 & 78.4\% & 1.417 \\
\midrule
\emph{LLM2feat}+AF      & 79.5\% & 1.221 & 49.0\% & \textbf{1.616} & 78.2\% & 1.420 \\
\emph{LLM2feat}+RF      & 85.3\% & 1.165 & 58.1\% & 1.862 & \textbf{84.7\%} & \textbf{1.286} \\
\emph{LLM2feat}+LLAMA  & \textbf{85.8\%} & \textbf{1.157} & 61.0\% & 1.773 & 84.2\% & 1.310 \\
\emph{LLM2feat}+AutoSK  & 85.7\% & \textbf{1.157} & \textbf{64.0\%} & 1.735 & 84.6\% & 1.310 \\
\bottomrule
\end{tabular}}
\end{table}

\subsection{Head-to-Head Comparison with \emph{trans2feat}}
\label{sec:trans2feat-comparison}

\citet{pellegrino2025transformer} release $20$ \emph{trans2feat} feature
variants (one per neural-network configuration) for CS and FLECC.
Table~\ref{tab:accuracy_loss_acc_test} compares only the \emph{best}
\emph{trans2feat} variant per toolchain. To examine the comparison more
carefully, we trained the same RF and AutoSK toolchains on every
\emph{trans2feat} variant, on every \emph{LLM2feat} run from $10$
independent agent syntheses, and on the canonical \emph{mzn2feat}
extractor. Table~\ref{tab:trans2feat_summary} reports the best, mean,
and worst test accuracy across each group. Even the \emph{worst}
\emph{LLM2feat} run beats the \emph{best} \emph{trans2feat} variant on
FLECC under RF, and clearly dominates on CS as well. Per-variant tables
for both toolchains and both problems are in
Appendix~\ref{appendix:transformer-detail}.

\begin{table}[htbp]
\caption{Head-to-head with \emph{trans2feat}: test-set $Acc$
aggregated over $10$ independent \emph{LLM2feat} runs and the $20$
released \emph{trans2feat} variants. \emph{mzn2feat} is a single
deterministic extractor. Best/Mean/Worst refer to the best/mean/worst
test accuracy across runs within each method; runs without a valid
selector (TO/F, Appendix~\ref{appendix:transformer-detail}) are
excluded, leaving $n=17$ \emph{trans2feat} variants and $n=9$
\emph{LLM2feat} runs on CS/AutoSK and $n=14$ variants on FLECC/AutoSK.}
\label{tab:trans2feat_summary}
\centering
\scalebox{0.90}{
\begin{tabular}{llccc}
\toprule
Problem & Toolchain & \emph{mzn2feat} & \emph{trans2feat} (Best / Mean / Worst) & \emph{LLM2feat} (Best / Mean / Worst) \\
\midrule
CS    & RF      & 54.7\% & 52.9\% / 43.9\% / 31.0\% & \textbf{58.3\%} / 57.8\% / 57.3\% \\
CS    & AutoSK  & 62.1\% & 54.5\% / 48.0\% / 35.8\% & \textbf{64.0\%} / 60.5\% / 52.8\% \\
FLECC & RF      & 76.4\% & 77.3\% / 51.1\% / 3.1\%  & \textbf{84.7\%} / 83.1\% / 81.6\% \\
FLECC & AutoSK  & 79.2\% & 78.4\% / 68.2\% / 8.3\%  & \textbf{84.6\%} / 83.6\% / 82.2\% \\
\bottomrule
\end{tabular}}
\end{table}

\subsection{Qualitative Feature Analysis}
\label{sec:qualitative}

To illustrate the solver-awareness of LLM-generated features, consider
VRP: the agent synthesizes features such as demand-distribution
statistics (\texttt{avg\_demand}) and depot-centrality metrics
(\texttt{depot\_centrality}). Such features enable the selector to
distinguish instances where MIP solvers excel from those better suited
to CP solvers~\citep{moreno2016experimental}, and they account for the
$+2.9$\,pp test-accuracy gain over \emph{mzn2feat} on VRP. Note that the
advantage cannot stem from feature-set size: \emph{LLM2feat} uses $50$
features against \emph{mzn2feat}'s $95$ and \emph{trans2feat}'s $116$.
However, the agent synthesizes \emph{relevant, complementary} feature
combinations that the curated list lacks, consistent with the lower
feature correlation (Q1) and the higher utilization (Q2) reported above.
We attribute the broader gains over classical baselines to three
properties of the generated extractors.

First, the extractors preserve structural information lost in flat
encodings. \emph{mzn2feat}~\citep{amadini14enhanced} flattens the
constraint model into a long list of primitive constraints, discarding
much of the high-level structure, whereas richer structural views often
dominate cheap flat features in
practice~\citep{dalla2023sat,shavit2024revisit}.

Second, the extractors materialize solver-aware quantities that flat
features miss. The LLM reads the model and \texttt{.dzn} files
(objective, global constraints, parameter roles) and the generated code
computes proxy quantities that solvers implicitly exploit, such as
tightness, propagation strength approximating domain-reduction ratios,
and supports per constraint family (e.g.\ \texttt{alldifferent},
\texttt{table}), along with distributional summaries
(mean/variance/skew).

Third, the agent adapts feature definitions to the problem family. For
VRP it instantiates domain priors as interpretable features such as
demand concentration, depot eccentricity, and route-length lower bounds
(e.g.\ MST surrogates), which separate MIP-friendly from CP-friendly
instances by leveraging the solvers'
strengths~\citep{moreno2016experimental}.

\subsection{Cost Analysis}
\label{sec:cost}

To quantify the cost of extractor synthesis, we ran the agent without
a wall-clock cap on four LLM backends, two commercial and two
open-weight (Table~\ref{tab:backend_overview}). As already noted in
Section~\ref{sec:settings}, o4-mini and Claude~Sonnet~4 produced a
working extractor in all $10$ of $10$ trials, while the open-weight
DeepSeek~R1 and Llama~3.3-8b-instruct failed in all $10$ trials because
they did not follow the check--fix--verify procedure. No
open-weight backend we tried sustained the protocol at the time of our
experiments; since the framework's \texttt{LM:} interface runs local
backends served via Ollama or LM~Studio (see supplement), stronger
future open-weight models can be plugged in directly, and we consider a
systematic open-weight evaluation worthwhile future work.

\begin{table}[htbp]
\begin{minipage}[t]{0.60\linewidth}
\centering
\captionsetup{width=\linewidth}
\caption{LLM backends evaluated as synthesis agents.}
\label{tab:backend_overview}
\scalebox{0.82}{
\begin{tabular}{lll}
\toprule
\textbf{Backend} & \textbf{Access} & \textbf{Success} \\
\midrule
OpenAI o4-mini-2025-04-16          & commercial {API}  & $10/10$ \\
Anthropic claude-sonnet-4-20250514 & commercial {API}  & $10/10$ \\
DeepSeek~R1                        & open weights      & $0/10$  \\
Llama~3.3-8b-instruct              & open weights      & $0/10$  \\
\bottomrule
\end{tabular}}
\end{minipage}\hfill
\begin{minipage}[t]{0.38\linewidth}
\centering
\captionsetup{width=\linewidth}
\caption{Statistics from $30$ \emph{LLM2feat} synthesis runs on o4-mini.}
\label{tab:cost_statistics}
\scalebox{0.82}{
\begin{tabular}{ll}
\toprule
\textbf{Metric} & \textbf{Value} \\
\midrule
Avg iterations    & $17.3$ \\
Avg wall time     & $210.0$\,s \\
Avg input tokens  & $259{,}913$ \\
Avg output tokens & $19{,}238$ \\
Avg total tokens  & $279{,}150$ \\
Avg cost          & \$$0.2695$ \\
\bottomrule
\end{tabular}}
\end{minipage}
\end{table}

Table~\ref{tab:cost_statistics} aggregates statistics over $30$
o4-mini runs on different problems until an extractor is generated:
the average single-extractor synthesis cost is $\approx 210$\,s of
wall-clock time and $\approx \$0.27$ in API charges, which is
negligible relative to running each solver for up to $20$ minutes per
instance. Once synthesized, an extractor can be reused indefinitely
across instances of the same problem family.

\section{Conclusion and Future Work}
\label{sec:conclusion}

We presented an LLM-based framework that synthesizes graph-theoretic,
interpretable feature extractors from symbolic constraint models
(MiniZinc) for algorithm selection. The agentic check--fix--verify
loop produces Python scripts that human experts can refine in a
``gray-box'' manner, and extractor synthesis is cheap enough to be
performed on demand. On three problem-specific benchmarks (VRP, CS,
FLECC) with a five-solver portfolio, our synthesized extractors yield
AS models that outperform both expert-curated \emph{mzn2feat} and
transformer-based \emph{trans2feat} baselines. Four concrete limits
define the current scope and the follow-up roadmap: the framework
produces one extractor per problem family rather than a universal one;
the agent's loop does not yet incorporate human feedback on the
generated code; the input formalism is restricted to MiniZinc; and
reliable synthesis currently requires strong commercial LLM backends,
as the open-weight models we evaluated failed the protocol
(Section~\ref{sec:cost}). Each of these is a concrete target for
follow-up work.

\section*{Impact Statement}

This work aims to improve algorithm selection (AS) for constraint
optimization by automating the creation of interpretable feature
extractors from high-level MiniZinc models. The positive impact is
reducing the manual feature-engineering burden that currently limits AS
to a small set of well-studied domains: auditable Python extractors let
practitioners build solver portfolios quickly for new problem classes,
with applications in scheduling, routing, and resource allocation.

The LLM-generated code may
carry errors or biases if deployed without validation. Both are
mitigated by the extractors' interpretability: the code is explicit, testable, and auditable by domain experts.

\paragraph{Acknowledgements.}
The authors acknowledge the support of the European Union's Horizon 2020 research and innovation
programme under the Maria Skłodowska-Curie grant agreement
No.~101034440\marginpar{\worldflag[width=18pt]{EU}}, the support of the
Austrian Science Fund (FWF), projects 10.55776/P36688, 10.55776/P36420, and
10.55776/COE12, and project PID2022-138506NB-C21 from Ministerio de Ciencia
e Innovación.

\bibliography{references}

\newpage
\appendix

\section{Hyperparameters}
\label{appendix:hyperparams}

Settings shared by Random Forest training and the LLAMA random-forest
classification mode are in Table~\ref{tab:rf_params}. AutoSklearn
settings are in Table~\ref{tab:autosklearn_params}.

\begin{table}[H]
\caption{Random Forest hyperparameters.}
\label{tab:rf_params}
\centering
\scalebox{0.85}{
\begin{tabular}{lll}
\toprule
\textbf{Parameter} & \textbf{Value} & \textbf{Description}\\
\midrule
n\_estimators & $300$ & More trees for complex constraint patterns\\
max\_depth & $20$ & Deeper trees capture CP relationships\\
min\_samples\_split & $5$ &\\
min\_samples\_leaf & $2$ &\\
resampling\_strategy & \texttt{cv} &\\
resampling\_strategy\_arguments & \texttt{\{folds: 5\}} &\\
max\_features & \texttt{sqrt} & Standard dimensionality reduction for tree diversity\\
class\_weight & \texttt{balanced} & Handles solver class imbalance\\
random\_state & $42$ &\\
\bottomrule
\end{tabular}
}
\end{table}

\begin{table}[H]
\caption{AutoSklearn standard configuration.}
\label{tab:autosklearn_params}
\centering
\begin{tabular}{ll}
\toprule
\textbf{Parameter} & \textbf{Value} \\
\midrule
time\_left\_for\_this\_task & User-specified ($300$--$3600$\,s) \\
per\_run\_time\_limit & time\_budget$\,/\,30$\,s \\
initial\_configurations\_via\_metalearning & $25$ \\
ensemble\_size & $50$ \\
resampling\_strategy & \texttt{cv} \\
resampling\_strategy\_arguments & \texttt{\{folds: 5\}} \\
scoring\_functions & \texttt{[accuracy, ranking]} \\
memory\_limit & $3072$\,MB \\
tmp\_folder & auto-generated \\
delete\_tmp\_folder\_after\_terminate & False \\
random\_state & $42$ \\
\bottomrule
\end{tabular}
\end{table}

\section{Additional Feature-Correlation Heatmaps}
\label{appendix:correlation-detail}

\begin{figure}[H]
  \centering
  \includegraphics[width=0.65\linewidth]{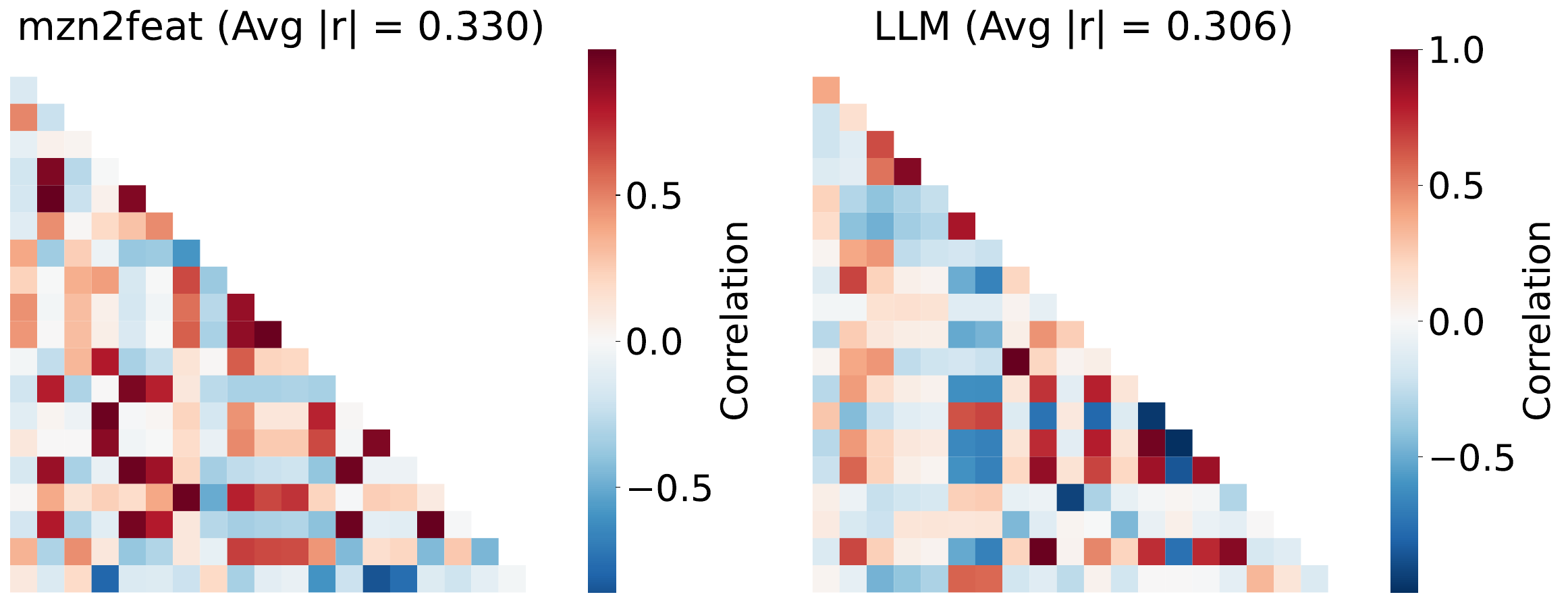}
  \caption{Feature correlation matrices for FLECC. \emph{LLM2feat}
  features have average absolute correlation $|r|=0.306$ vs.\
  $|r|=0.330$ for \emph{mzn2feat}.}
  \label{fig:flecc_correlation}
\end{figure}

\begin{figure}[H]
  \centering
  \includegraphics[width=0.65\linewidth]{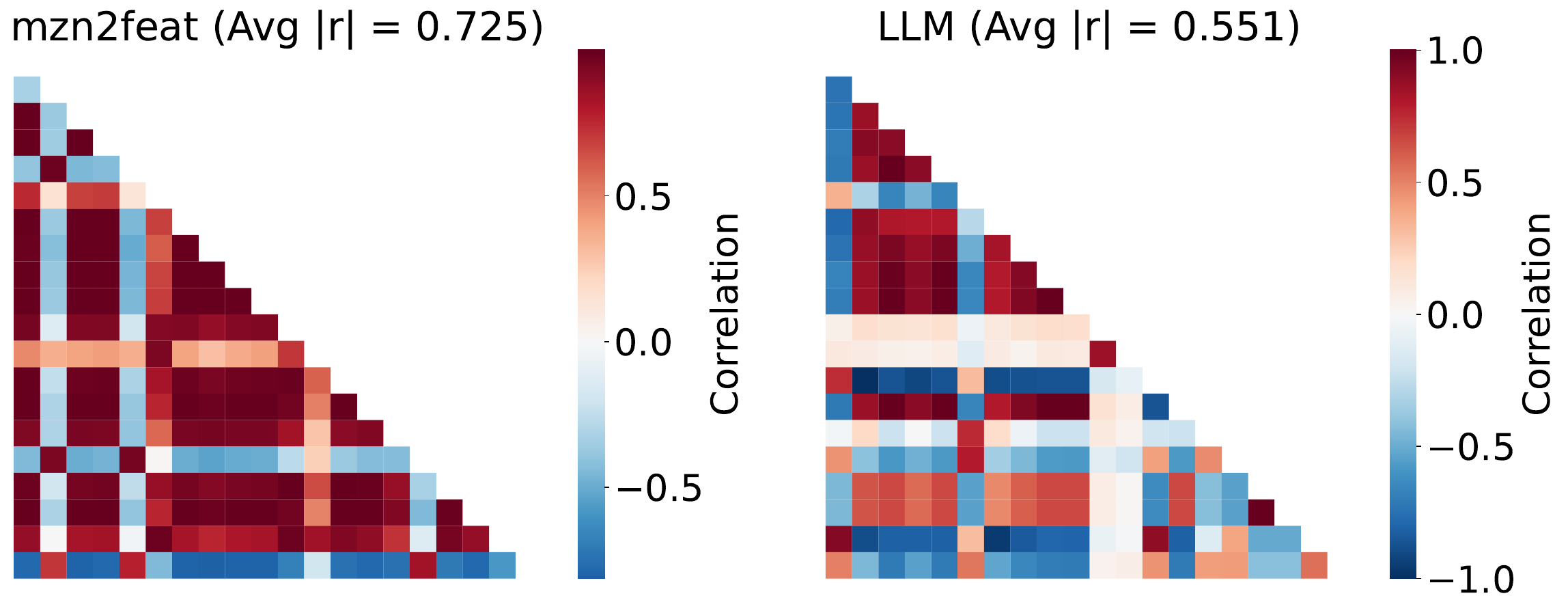}
  \caption{Feature correlation matrices for CS. \emph{LLM2feat} features
  have $|r|=0.551$ vs.\ $|r|=0.725$ for \emph{mzn2feat}, a $24\%$
  improvement in diversity.}
  \label{fig:car_correlation}
\end{figure}

Figures~\ref{fig:flecc_correlation}--\ref{fig:car_correlation} show the
correlation matrices for the FLECC and CS problems.

\section{LLM Model Selection and Sensitivity}
\label{appendix:llm-sensitivity}

We evaluated several LLMs as the agent backend. Strong agentic models
(o4-mini, Claude Sonnet) reliably followed the check--fix--verify
protocol; smaller open-source models (Llama~3.3, DeepSeek~R1) failed
under our prompts (see Section~\ref{sec:cost}). For the
problem-specific framework we used OpenAI o4-mini-2025-04-16 and ran $10$
independent syntheses per problem. Tables~\ref{tab:flecc_rf}--\ref{tab:vrp_autosk}
list the three highest-scoring runs per problem and toolchain alongside
\emph{mzn2feat} baselines; full per-variant tables (10 LLM2feat runs
vs.\ 20 \emph{trans2feat} variants) are in
Appendix~\ref{appendix:transformer-detail}. Across the top-three runs
shown here, test accuracy fluctuates within roughly $3$\,pp and every
shown LLM2feat run outperforms the corresponding \emph{mzn2feat}
baseline. The full $10$-run distribution is wider on CS/AutoSK in
particular, where two syntheses produced under-fitting extractors
(test accuracy $\approx 0.528$, with correspondingly low training
accuracy); the formal significance analysis on the full distributions
appears in Section~\ref{sec:trans2feat-comparison}.

\paragraph{Training-set and ranking-loss results}
Tables~\ref{tab:training_acc_loss}--\ref{tab:training_rank_loss} report
the results omitted from the main text: training-set accuracy under
$Acc$ loss; test-set and training-set accuracy/ranking under $Rank$
loss.

\begin{table}[H]
\caption{Training-set accuracy/ranking with $Acc$ as loss
(problem-specific toolchains).}
\label{tab:training_acc_loss}
\centering
\scalebox{0.95}{
\begin{tabular}{lcccccc}
\toprule
& \multicolumn{2}{c}{VRP} & \multicolumn{2}{c}{CS} & \multicolumn{2}{c}{FLECC} \\
\cmidrule(lr){2-3} \cmidrule(lr){4-5} \cmidrule(lr){6-7}
                  & Acc & Rank & Acc & Rank & Acc & Rank \\
\midrule
SB                & 78.7\% & 1.237 & 49.4\% & 1.601 & 80.6\% & 1.377 \\
\emph{mzn2feat}+AF      & 78.7\% & 1.237 & 49.4\% & 1.601 & 80.6\% & 1.377 \\
\emph{mzn2feat}+RF      & 94.7\% & 1.057 & 85.7\% & 1.339 & 95.9\% & 1.063 \\
\emph{mzn2feat}+LLAMA  & 92.0\% & 1.089 & 87.5\% & 1.248 & 91.5\% & 1.167 \\
\emph{mzn2feat}+AutoSK  & 87.2\% & 1.138 & 65.0\% & 1.723 & 85.0\% & 1.295 \\
\emph{LLM2feat}+RF      & 98.5\% & 1.016 & 90.6\% & 1.202 & 98.1\% & 1.029 \\
\emph{LLM2feat}+LLAMA  & 97.8\% & 1.025 & 89.9\% & 1.164 & 98.3\% & 1.039 \\
\emph{LLM2feat}+AutoSK  & 88.0\% & 1.131 & 67.1\% & 1.646 & 95.1\% & 1.104 \\
\bottomrule
\end{tabular}}
\end{table}

\begin{table}[H]
\caption{Test-set accuracy/ranking with $Rank$ as loss.}
\label{tab:test_rank_loss}
\centering
\begin{tabular}{lcccccc}
\toprule
& \multicolumn{2}{c}{VRP} & \multicolumn{2}{c}{CS} & \multicolumn{2}{c}{FLECC} \\
\cmidrule(lr){2-3} \cmidrule(lr){4-5} \cmidrule(lr){6-7}
                   & Acc & Rank & Acc & Rank & Acc & Rank \\
\midrule
SB                 & 79.5\% & 1.221 & 49.0\% & 1.616 & 78.2\% & 1.420 \\
\emph{mzn2feat}+RF       & 83.0\% & 1.184 & 58.5\% & 1.680 & 79.5\% & 1.391 \\
\emph{mzn2feat}+AutoSK   & 84.4\% & 1.166 & 62.1\% & 1.787 & 79.4\% & 1.394 \\
\emph{LLM2feat}+RF       & 85.7\% & 1.155 & 61.3\% & 1.681 & 83.5\% & 1.338 \\
\emph{LLM2feat}+AutoSK   & 85.4\% & 1.160 & 64.0\% & 1.738 & 84.6\% & 1.310 \\
\bottomrule
\end{tabular}
\end{table}

\begin{table}[H]
\caption{Training-set accuracy/ranking with $Rank$ as loss.}
\label{tab:training_rank_loss}
\centering
\begin{tabular}{lcccccc}
\toprule
& \multicolumn{2}{c}{VRP} & \multicolumn{2}{c}{CS} & \multicolumn{2}{c}{FLECC} \\
\cmidrule(lr){2-3} \cmidrule(lr){4-5} \cmidrule(lr){6-7}
                   & Acc & Rank & Acc & Rank & Acc & Rank \\
\midrule
SB                 & 78.7\% & 1.237 & 49.4\% & 1.601 & 80.6\% & 1.377 \\
\emph{mzn2feat}+RF       & 91.7\% & 1.091 & 83.5\% & 1.231 & 88.3\% & 1.232 \\
\emph{mzn2feat}+AutoSK   & 87.2\% & 1.138 & 65.0\% & 1.723 & 85.0\% & 1.296 \\
\emph{LLM2feat}+RF       & 97.1\% & 1.032 & 87.9\% & 1.164 & 95.9\% & 1.083 \\
\emph{LLM2feat}+AutoSK   & 87.8\% & 1.133 & 68.9\% & 1.618 & 95.1\% & 1.103 \\
\bottomrule
\end{tabular}
\end{table}

\paragraph{Per-extractor sensitivity tables}

\begin{table}[H]
\centering
\caption{\emph{LLM2feat}+RF performance for FLECC.}
\label{tab:flecc_rf}
\begin{tabular}{llcccc}
\toprule
Extractor & Loss & Train Acc & Test Acc & Train Rank & Test Rank \\
\midrule
\multicolumn{2}{l}{\textbf{Single Best (gecode)}} & 0.806 & 0.782 & 1.377 & 1.420 \\
\midrule
\emph{mzn2feat} & accuracy & 0.959 & 0.764 & 1.063 & 1.426 \\
\emph{mzn2feat} & ranking  & 0.883 & 0.795 & 1.232 & 1.391 \\
\midrule
LLM-20250908123730 & accuracy & 0.981 & 0.847 & 1.029 & 1.286 \\
LLM-20250908123925 & accuracy & 0.996 & 0.818 & 1.008 & 1.353 \\
LLM-20250908124149 & accuracy & 0.994 & 0.836 & 1.011 & 1.315 \\
LLM-20250908123730 & ranking  & 0.959 & 0.835 & 1.083 & 1.338 \\
LLM-20250908123925 & ranking  & 0.935 & 0.809 & 1.121 & 1.384 \\
LLM-20250908124149 & ranking  & 0.952 & 0.826 & 1.095 & 1.358 \\
\bottomrule
\end{tabular}
\end{table}

\begin{table}[H]
\centering
\caption{\emph{LLM2feat}+RF performance for CS.}
\label{tab:cs_rf}
\begin{tabular}{llcccc}
\toprule
Extractor & Loss & Train Acc & Test Acc & Train Rank & Test Rank \\
\midrule
\multicolumn{2}{l}{\textbf{Single Best (cplex)}} & 0.494 & 0.490 & 1.601 & 1.616 \\
\midrule
\emph{mzn2feat} & accuracy & 0.857 & 0.547 & 1.339 & 1.943 \\
\emph{mzn2feat} & ranking  & 0.835 & 0.585 & 1.231 & 1.680 \\
\midrule
LLM-20250908123608 & accuracy & 0.903 & 0.577 & 1.193 & 1.874 \\
LLM-20250908123905 & accuracy & 0.913 & 0.578 & 1.181 & 1.861 \\
LLM-20250908124041 & accuracy & 0.906 & 0.581 & 1.202 & 1.862 \\
LLM-20250908123608 & ranking  & 0.868 & 0.607 & 1.179 & 1.700 \\
LLM-20250908123905 & ranking  & 0.879 & 0.613 & 1.165 & 1.681 \\
LLM-20250908124041 & ranking  & 0.875 & 0.606 & 1.171 & 1.704 \\
\bottomrule
\end{tabular}
\end{table}

\begin{table}[H]
\centering
\caption{\emph{LLM2feat}+RF performance for VRP.}
\label{tab:vrp_rf}
\begin{tabular}{llcccc}
\toprule
Extractor & Loss & Train Acc & Test Acc & Train Rank & Test Rank \\
\midrule
\multicolumn{2}{l}{\textbf{Single Best (scip)}} & 0.787 & 0.795 & 1.237 & 1.221 \\
\midrule
\emph{mzn2feat} & accuracy & 0.947 & 0.824 & 1.057 & 1.195 \\
\emph{mzn2feat} & ranking  & 0.916 & 0.830 & 1.091 & 1.184 \\
\midrule
LLM-20250908115627 & accuracy & 0.993 & 0.850 & 1.008 & 1.169 \\
LLM-20250908121942 & accuracy & 0.994 & 0.848 & 1.007 & 1.170 \\
LLM-20250908123205 & accuracy & 0.985 & 0.853 & 1.016 & 1.165 \\
LLM-20250908115627 & ranking  & 0.971 & 0.852 & 1.032 & 1.163 \\
LLM-20250908121942 & ranking  & 0.972 & 0.857 & 1.032 & 1.155 \\
LLM-20250908123205 & ranking  & 0.970 & 0.852 & 1.034 & 1.163 \\
\bottomrule
\end{tabular}
\end{table}

\begin{table}[H]
\centering
\caption{\emph{LLM2feat}+AutoSK performance for FLECC.}
\label{tab:flecc_autosk}
\begin{tabular}{llcccc}
\toprule
Extractor & Loss & Train Acc & Test Acc & Train Rank & Test Rank \\
\midrule
\multicolumn{2}{l}{\textbf{Single Best (gecode)}} & 0.806 & 0.782 & 1.377 & 1.420 \\
\midrule
\emph{mzn2feat} & accuracy & 0.850 & 0.792 & 1.295 & 1.396 \\
\emph{mzn2feat} & ranking  & 0.850 & 0.794 & 1.296 & 1.394 \\
\midrule
LLM-20250908124149 & accuracy & 0.952 & 0.844 & 1.101 & 1.308 \\
LLM-20250908123730 & accuracy & 0.951 & 0.846 & 1.104 & 1.310 \\
LLM-20250908123925 & accuracy & 0.908 & 0.831 & 1.187 & 1.342 \\
LLM-20250908124149 & ranking  & 0.951 & 0.845 & 1.103 & 1.308 \\
LLM-20250908123730 & ranking  & 0.951 & 0.846 & 1.104 & 1.310 \\
LLM-20250908123925 & ranking  & 0.911 & 0.836 & 1.183 & 1.322 \\
\bottomrule
\end{tabular}
\end{table}

\begin{table}[H]
\centering
\caption{\emph{LLM2feat}+AutoSK performance for CS.}
\label{tab:cs_autosk}
\begin{tabular}{llcccc}
\toprule
Extractor & Loss & Train Acc & Test Acc & Train Rank & Test Rank \\
\midrule
\multicolumn{2}{l}{\textbf{Single Best (cplex)}} & 0.494 & 0.490 & 1.601 & 1.616 \\
\midrule
\emph{mzn2feat} & accuracy & 0.650 & 0.621 & 1.723 & 1.787 \\
\emph{mzn2feat} & ranking  & 0.650 & 0.621 & 1.723 & 1.787 \\
\midrule
LLM-20250908124041 & accuracy & 0.671 & 0.640 & 1.646 & 1.735 \\
LLM-20250908123905 & accuracy & 0.679 & 0.634 & 1.667 & 1.781 \\
LLM-20250908123608 & accuracy & 0.686 & 0.635 & 1.652 & 1.782 \\
LLM-20250908124041 & ranking  & 0.670 & 0.640 & 1.655 & 1.738 \\
LLM-20250908123905 & ranking  & 0.675 & 0.638 & 1.655 & 1.766 \\
LLM-20250908123608 & ranking  & 0.689 & 0.639 & 1.618 & 1.743 \\
\bottomrule
\end{tabular}
\end{table}

\begin{table}[H]
\centering
\caption{\emph{LLM2feat}+AutoSK performance for VRP.}
\label{tab:vrp_autosk}
\begin{tabular}{llcccc}
\toprule
Extractor & Loss & Train Acc & Test Acc & Train Rank & Test Rank \\
\midrule
\multicolumn{2}{l}{\textbf{Single Best (scip)}} & 0.787 & 0.795 & 1.237 & 1.221 \\
\midrule
\emph{mzn2feat} & accuracy & 0.872 & 0.844 & 1.138 & 1.166 \\
\emph{mzn2feat} & ranking  & 0.872 & 0.844 & 1.138 & 1.166 \\
\midrule
LLM-20250908121942 & accuracy & 0.851 & 0.852 & 1.161 & 1.162 \\
LLM-20250908123205 & accuracy & 0.880 & 0.857 & 1.131 & 1.157 \\
LLM-20250908115627 & accuracy & 0.872 & 0.854 & 1.138 & 1.160 \\
LLM-20250908121942 & ranking  & 0.851 & 0.852 & 1.161 & 1.162 \\
LLM-20250908123205 & ranking  & 0.878 & 0.854 & 1.133 & 1.160 \\
LLM-20250908115627 & ranking  & 0.859 & 0.850 & 1.152 & 1.162 \\
\bottomrule
\end{tabular}
\end{table}

\section{Detailed Transformer Comparison}
\label{appendix:transformer-detail}

We obtain $20$ \emph{trans2feat} feature sets from the official
repository\footnote{\url{https://github.com/SeppiaBrilla/EFE_project/tree/master/data/features}}
(one per neural-network variant in~\citet{pellegrino2025transformer}) and
train both RF and AutoSK toolchains on each. Tables
\ref{tab:cs_rf_full}--\ref{tab:flecc_autosk_full} give the full per-variant
results for CS and FLECC. ``TO'' marks runs that exceeded the
AutoSklearn $1800$\,s training budget~\citep{feurer20Autosklearn} without
producing a model; ``F'' marks runs whose AutoSklearn fit collapsed to a
degenerate constant-class model (test accuracy ${<}1\%$). Both are
excluded from all aggregate statistics and significance tests; the raw
result files for the F runs ship with the supplement.

\begin{table}[H]
\centering
\caption{Car Sequencing: Random Forest with accuracy loss.}
\label{tab:cs_rf_full}
\scalebox{0.85}{
\begin{tabular}{llcccccc}
\toprule
\textbf{Type} & \textbf{Extractor} & \textbf{Features} & \multicolumn{2}{c}{\textbf{Accuracy}} & \multicolumn{2}{c}{\textbf{Avg Rank}} \\
\cmidrule(lr){4-5} \cmidrule(lr){6-7}
 & & & Train & Test & Train & Test \\
\midrule
\emph{LLM2feat}    & LLM2feat-1  & $50$ & 0.909 & 0.583 & 1.187 & 1.857 \\
                & LLM2feat-2  & $50$ & 0.906 & 0.581 & 1.202 & 1.862 \\
                & LLM2feat-3  & $50$ & 0.920 & 0.579 & 1.170 & 1.855 \\
                & LLM2feat-4  & $50$ & 0.908 & 0.579 & 1.188 & 1.867 \\
                & LLM2feat-5  & $50$ & 0.913 & 0.578 & 1.181 & 1.861 \\
                & LLM2feat-6  & $50$ & 0.899 & 0.578 & 1.213 & 1.867 \\
                & LLM2feat-7  & $50$ & 0.903 & 0.577 & 1.193 & 1.874 \\
                & LLM2feat-8  & $50$ & 0.915 & 0.577 & 1.180 & 1.872 \\
                & LLM2feat-9  & $50$ & 0.899 & 0.575 & 1.197 & 1.876 \\
                & LLM2feat-10 & $50$ & 0.904 & 0.573 & 1.193 & 1.875 \\
\midrule
\emph{mzn2feat}    & mzn2feat    & $95$ & 0.857 & 0.547 & 1.339 & 1.943 \\
\midrule
\emph{trans2feat} & com-6     & $116$ & 0.869 & 0.529 & 1.344 & 2.004 \\
                & com-9     & $116$ & 0.833 & 0.529 & 1.425 & 1.982 \\
                & com-3     & $116$ & 0.861 & 0.521 & 1.384 & 2.030 \\
                & com-2     & $116$ & 0.850 & 0.518 & 1.398 & 2.039 \\
                & com-1     & $116$ & 0.840 & 0.512 & 1.422 & 2.048 \\
                & com-5     & $116$ & 0.838 & 0.508 & 1.420 & 2.044 \\
                & com-0     & $116$ & 0.857 & 0.502 & 1.391 & 2.081 \\
                & com-7     & $116$ & 0.843 & 0.502 & 1.415 & 2.070 \\
                & com-8     & $116$ & 0.773 & 0.501 & 1.569 & 2.083 \\
                & com-4     & $116$ & 0.826 & 0.496 & 1.438 & 2.088 \\
                & neural-7  & $116$ & 0.789 & 0.482 & 1.522 & 2.110 \\
                & neural-6  & $116$ & 0.558 & 0.418 & 2.394 & 2.637 \\
                & neural-0  & $116$ & 0.557 & 0.411 & 2.371 & 2.627 \\
                & neural-4  & $116$ & 0.472 & 0.383 & 2.601 & 2.758 \\
                & neural-8  & $116$ & 0.428 & 0.368 & 2.723 & 2.836 \\
                & neural-2  & $116$ & 0.343 & 0.333 & 2.861 & 2.886 \\
                & neural-9  & $116$ & 0.366 & 0.326 & 2.735 & 2.828 \\
                & neural-3  & $116$ & 0.329 & 0.315 & 2.975 & 3.026 \\
                & neural-5  & $116$ & 0.309 & 0.314 & 2.978 & 2.985 \\
                & neural-1  & $116$ & 0.319 & 0.310 & 3.009 & 3.039 \\
\bottomrule
\end{tabular}}
\end{table}

\begin{table}[H]
\centering
\caption{FLECC: Random Forest with accuracy loss.}
\label{tab:flecc_rf_full}
\scalebox{0.85}{
\begin{tabular}{llcccccc}
\toprule
\textbf{Type} & \textbf{Extractor} & \textbf{Features} & \multicolumn{2}{c}{\textbf{Accuracy}} & \multicolumn{2}{c}{\textbf{Avg Rank}} \\
\cmidrule(lr){4-5} \cmidrule(lr){6-7}
 & & & Train & Test & Train & Test \\
\midrule
\emph{LLM2feat}    & LLM2feat-1  & $50$ & 0.981 & 0.847 & 1.029 & 1.286 \\
                & LLM2feat-2  & $50$ & 0.989 & 0.842 & 1.018 & 1.310 \\
                & LLM2feat-3  & $50$ & 0.984 & 0.842 & 1.025 & 1.305 \\
                & LLM2feat-4  & $50$ & 0.994 & 0.836 & 1.011 & 1.315 \\
                & LLM2feat-5  & $50$ & 0.995 & 0.835 & 1.009 & 1.325 \\
                & LLM2feat-6  & $50$ & 0.980 & 0.831 & 1.033 & 1.324 \\
                & LLM2feat-7  & $50$ & 0.989 & 0.826 & 1.019 & 1.334 \\
                & LLM2feat-8  & $50$ & 0.996 & 0.818 & 1.008 & 1.353 \\
                & LLM2feat-9  & $50$ & 0.994 & 0.818 & 1.012 & 1.360 \\
                & LLM2feat-10 & $50$ & 0.976 & 0.816 & 1.041 & 1.350 \\
\midrule
\emph{mzn2feat}    & mzn2feat    & $95$ & 0.959 & 0.764 & 1.063 & 1.426 \\
\midrule
\emph{trans2feat} & com-3     & $116$ & 0.803 & 0.773 & 1.389 & 1.447 \\
                & com-2     & $116$ & 0.800 & 0.772 & 1.399 & 1.459 \\
                & com-1     & $116$ & 0.765 & 0.746 & 1.532 & 1.552 \\
                & neural-9  & $116$ & 0.958 & 0.745 & 1.067 & 1.480 \\
                & neural-7  & $116$ & 0.958 & 0.743 & 1.072 & 1.490 \\
                & com-8     & $116$ & 0.941 & 0.734 & 1.107 & 1.517 \\
                & com-7     & $116$ & 0.949 & 0.732 & 1.085 & 1.501 \\
                & com-9     & $116$ & 0.933 & 0.731 & 1.108 & 1.510 \\
                & neural-8  & $116$ & 0.947 & 0.727 & 1.088 & 1.522 \\
                & com-4     & $116$ & 0.727 & 0.708 & 1.637 & 1.660 \\
                & com-5     & $116$ & 0.688 & 0.673 & 1.775 & 1.785 \\
                & neural-1  & $116$ & 0.688 & 0.669 & 1.805 & 1.825 \\
                & neural-0  & $116$ & 0.612 & 0.590 & 2.103 & 2.136 \\
                & com-6     & $116$ & 0.472 & 0.444 & 2.172 & 2.267 \\
                & neural-6  & $116$ & 0.095 & 0.114 & 2.466 & 2.455 \\
                & neural-3  & $116$ & 0.095 & 0.111 & 2.510 & 2.496 \\
                & neural-2  & $116$ & 0.086 & 0.108 & 2.610 & 2.573 \\
                & com-0     & $116$ & 0.030 & 0.033 & 3.933 & 3.968 \\
                & neural-5  & $116$ & 0.027 & 0.031 & 3.957 & 3.985 \\
                & neural-4  & $116$ & 0.027 & 0.031 & 3.957 & 3.985 \\
\bottomrule
\end{tabular}}
\end{table}

\begin{table}[H]
\centering
\caption{Car Sequencing: AutoSklearn with accuracy loss.
F~$=$~degenerate failed fit (test accuracy ${<}1\%$), excluded from all
aggregates.}
\label{tab:cs_autosk_full}
\scalebox{0.85}{
\begin{tabular}{llcccccc}
\toprule
\textbf{Type} & \textbf{Extractor} & \textbf{Features} & \multicolumn{2}{c}{\textbf{Accuracy}} & \multicolumn{2}{c}{\textbf{Avg Rank}} \\
\cmidrule(lr){4-5} \cmidrule(lr){6-7}
 & & & Train & Test & Train & Test \\
\midrule
\emph{LLM2feat}    & LLM2feat-1  & $50$ & 0.671 & 0.640 & 1.646 & 1.735 \\
                & LLM2feat-2  & $50$ & 0.686 & 0.635 & 1.652 & 1.782 \\
                & LLM2feat-3  & $50$ & 0.679 & 0.634 & 1.667 & 1.781 \\
                & LLM2feat-4  & $50$ & 0.709 & 0.630 & 1.578 & 1.756 \\
                & LLM2feat-5  & $50$ & 0.669 & 0.629 & 1.631 & 1.729 \\
                & LLM2feat-6  & $50$ & 0.641 & 0.616 & 1.707 & 1.788 \\
                & LLM2feat-7  & $50$ & 0.651 & 0.607 & 1.696 & 1.802 \\
                & LLM2feat-8  & $50$ & 0.660 & 0.529 & 1.738 & 2.011 \\
                & LLM2feat-9  & $50$ & 0.658 & 0.528 & 1.754 & 2.017 \\
                & LLM2feat-10 (F) & $50$ & --- & --- & --- & --- \\
\midrule
\emph{mzn2feat}    & mzn2feat    & $95$ & 0.650 & 0.621 & 1.723 & 1.787 \\
\midrule
\emph{trans2feat} & com-4     & $116$ & 0.867 & 0.545 & 1.271 & 1.931 \\
                & com-7     & $116$ & 0.869 & 0.542 & 1.270 & 1.927 \\
                & neural-9  & $116$ & 0.568 & 0.512 & 2.142 & 2.248 \\
                & neural-8  & $116$ & 0.521 & 0.499 & 2.435 & 2.492 \\
                & neural-1  & $116$ & 0.521 & 0.498 & 2.429 & 2.493 \\
                & neural-6  & $116$ & 0.518 & 0.495 & 2.420 & 2.482 \\
                & neural-3  & $116$ & 0.519 & 0.492 & 2.453 & 2.522 \\
                & neural-5  & $116$ & 0.511 & 0.491 & 2.493 & 2.553 \\
                & neural-2  & $116$ & 0.534 & 0.491 & 2.405 & 2.500 \\
                & com-6     & $116$ & 0.582 & 0.483 & 1.971 & 2.164 \\
                & com-9     & $116$ & 0.570 & 0.468 & 1.998 & 2.183 \\
                & neural-4  & $116$ & 0.558 & 0.465 & 2.419 & 2.570 \\
                & com-2     & $116$ & 0.570 & 0.462 & 2.015 & 2.224 \\
                & com-5     & $116$ & 0.545 & 0.456 & 2.063 & 2.236 \\
                & com-8     & $116$ & 0.523 & 0.452 & 2.133 & 2.270 \\
                & neural-7  & $116$ & 0.535 & 0.445 & 2.088 & 2.256 \\
                & neural-0  & $116$ & 0.397 & 0.358 & 2.709 & 2.798 \\
                & com-0 (F) & $116$ & --- & --- & --- & --- \\
                & com-1 (F) & $116$ & --- & --- & --- & --- \\
                & com-3 (F) & $116$ & --- & --- & --- & --- \\
\bottomrule
\end{tabular}}
\end{table}

\begin{table}[H]
\centering
\caption{FLECC: AutoSklearn with accuracy loss. TO~$=$~$1800$\,s timeout
without a model; F~$=$~degenerate failed fit.}
\label{tab:flecc_autosk_full}
\scalebox{0.85}{
\begin{tabular}{llcccccc}
\toprule
\textbf{Type} & \textbf{Extractor} & \textbf{Features} & \multicolumn{2}{c}{\textbf{Accuracy}} & \multicolumn{2}{c}{\textbf{Avg Rank}} \\
\cmidrule(lr){4-5} \cmidrule(lr){6-7}
 & & & Train & Test & Train & Test \\
\midrule
\emph{LLM2feat}    & LLM2feat-1  & $50$ & 0.951 & 0.846 & 1.104 & 1.310 \\
                & LLM2feat-2  & $50$ & 0.946 & 0.845 & 1.113 & 1.311 \\
                & LLM2feat-3  & $50$ & 0.952 & 0.844 & 1.101 & 1.308 \\
                & LLM2feat-4  & $50$ & 0.886 & 0.838 & 1.231 & 1.325 \\
                & LLM2feat-5  & $50$ & 0.924 & 0.837 & 1.157 & 1.324 \\
                & LLM2feat-6  & $50$ & 0.904 & 0.836 & 1.195 & 1.330 \\
                & LLM2feat-7  & $50$ & 0.908 & 0.831 & 1.187 & 1.342 \\
                & LLM2feat-8  & $50$ & 0.897 & 0.830 & 1.209 & 1.342 \\
                & LLM2feat-9  & $50$ & 0.872 & 0.828 & 1.250 & 1.340 \\
                & LLM2feat-10 & $50$ & 0.919 & 0.822 & 1.164 & 1.356 \\
\midrule
\emph{mzn2feat}    & mzn2feat    & $95$ & 0.850 & 0.792 & 1.295 & 1.396 \\
\midrule
\emph{trans2feat} & com-7     & $116$ & 0.819 & 0.784 & 1.357 & 1.417 \\
                & com-8     & $116$ & 0.809 & 0.783 & 1.371 & 1.418 \\
                & neural-9  & $116$ & 0.816 & 0.783 & 1.361 & 1.417 \\
                & com-4     & $116$ & 0.807 & 0.783 & 1.375 & 1.419 \\
                & neural-8  & $116$ & 0.826 & 0.783 & 1.345 & 1.417 \\
                & com-2     & $116$ & 0.806 & 0.782 & 1.377 & 1.420 \\
                & neural-1  & $116$ & 0.806 & 0.782 & 1.377 & 1.420 \\
                & neural-2  & $116$ & 0.806 & 0.782 & 1.377 & 1.420 \\
                & neural-6  & $116$ & 0.806 & 0.782 & 1.377 & 1.420 \\
                & com-6     & $116$ & 0.807 & 0.782 & 1.374 & 1.420 \\
                & com-9     & $116$ & 0.807 & 0.781 & 1.374 & 1.422 \\
                & neural-7  & $116$ & 0.806 & 0.781 & 1.376 & 1.422 \\
                & neural-4  & $116$ & 0.082 & 0.083 & 2.830 & 2.817 \\
                & neural-5  & $116$ & 0.082 & 0.083 & 2.830 & 2.817 \\
                & neural-3 (TO) & $116$ & --- & --- & --- & --- \\
                & com-0 (TO) & $116$ & --- & --- & --- & --- \\
                & com-1 (TO) & $116$ & --- & --- & --- & --- \\
                & com-3 (TO) & $116$ & --- & --- & --- & --- \\
                & com-5 (TO) & $116$ & --- & --- & --- & --- \\
                & neural-0 (TO) & $116$ & --- & --- & --- & --- \\
\bottomrule
\end{tabular}}
\end{table}

\end{document}